\documentclass[10pt,twocolumn,letterpaper]{article}

\usepackage{cvpr}
\usepackage{times}
\usepackage{epsfig}
\usepackage{graphicx}
\usepackage{amsmath}
\DeclareMathOperator*{\argmax}{argmax}
\usepackage{amssymb}

\usepackage[pagebackref=true,breaklinks=true,letterpaper=true,colorlinks,bookmarks=false]{hyperref}

\cvprfinalcopy % *** Uncomment this line for the final submission

\def\cvprPaperID{14} % *** Enter the CVPR Paper ID here

\ifcvprfinal\fi
\begin{document}

%%%%%%%%% TITLE
\title{Enabling Computer Vision Driven Assistive Devices for the Visually Impaired via Micro-architecture Design Exploration\thanks{We thank NSERC, Canada Research Chairs program, and
Microsoft.}}

\author{Linda Wang\\
University of Waterloo\\
Waterloo ON N2L 3G1, Canada\\
{\tt\small linda.wang@uwaterloo.ca}
\and
Alexander Wong\\
Waterloo AI Institute, University of Waterloo\\
Waterloo ON N2L 3G1, Canada\\
{\tt\small a28wong@uwaterloo.ca}
}

\maketitle
%\thispagestyle{empty}

%%%%%%%%% ABSTRACT
\begin{abstract}
   Recent improvements in object detection have shown potential to aid in tasks where previous solutions were not able to achieve. A particular area is assistive devices for individuals with visual impairment. While state-of-the-art deep neural networks have been shown to achieve superior object detection performance, their high computational and memory requirements make them cost prohibitive for on-device operation. Alternatively, cloud-based operation leads to privacy concerns, both not attractive to potential users. To address these challenges, this study investigates creating an efficient object detection network specifically for OLIV, an AI-powered assistant for object localization for the visually impaired, via micro-architecture design exploration. In particular, we formulate the problem of finding an optimal network micro-architecture as an numerical optimization problem, where we find the set of hyperparameters controlling the MobileNetV2-SSD network micro-architecture that maximizes a modified NetScore objective function for the MSCOCO-OLIV dataset of indoor objects. Experimental results show that such a micro-architecture design exploration strategy leads to a compact deep neural network with a balanced trade-off between accuracy, size, and speed, making it well-suited for enabling on-device computer vision driven assistive devices for the visually impaired.
\end{abstract}

%%%%%%%%% BODY TEXT
\section{Introduction}
\vspace{-0.06in}
The success of AlexNet on the 2012 ImageNet challenge demonstrated the efficacy of convolution neural networks for image recognition~\cite{alexnet}. The realization of convolution neural networks when applied to large datasets led to rapid advancements in not only image recognition, but also other areas of computer vision, such as depth estimation and object detection. As computer vision continues to improve, there is potential to aid in tasks where previous solutions struggled. One area where computer vision can make an impact is assisting those with visual impairment. According to \cite{who_2018}, there are 36 million people in the world who are legally blind. For these individuals, undertaking daily tasks are demanding. Recent advances in computer vision have the potential to assist them in their daily tasks, such as providing scene descriptions, identifying people, and locating displaced objects.

An integral part of many assistive devices for the visually impaired is object detection. However many object detection neural networks are large and require powerful graphical processing units (GPUs), leading to a trade-off between cost and privacy. If the device contains a local GPU to protect autonomy over personal data, affordability decreases. As an alternative, cloud services provide a more affordable solution at scale, however, privacy concerns arise over data access. In addition, the increasing complexity of these deep neural networks is one of the obstacles to widespread deployment on edge devices, where computational power and memory capacity are significantly lower than GPUs.

To address these challenges, we propose OLIV, an AI-powered assistant for object localization for the visually impaired~\cite{Wang_OLIV_2018}. For the system to be cost efficient and reduce privacy concerns, the object detection aspect of OLIV must be able to run locally. To achieve this, a network architecture exploration approach is investigated to create an optimal object detection network. In particular, this can be formulated as a numerical optimization problem, where we find a set of hyperparameters that optimizes the trade-offs between accuracy, speed, and size of the deep neural network.

\section{OLIV}
\begin{figure*}
\begin{center}
\vspace{-0.3in}
\includegraphics[width=\textwidth]{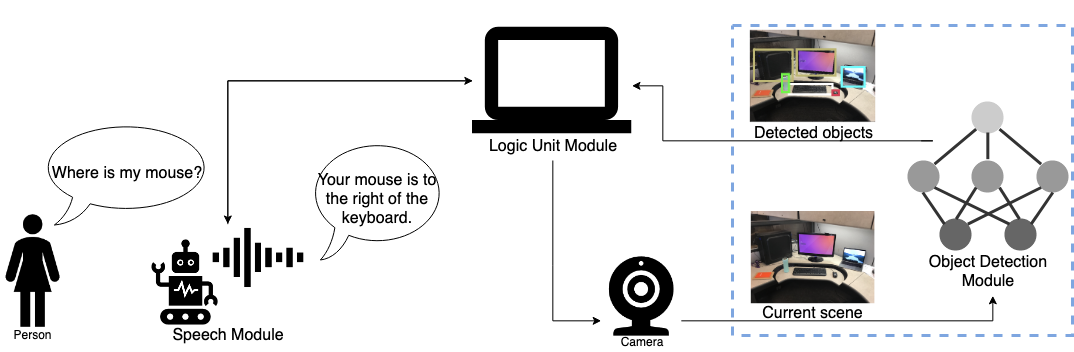}
\end{center}
   \caption{System overview of OLIV, which consists of three main components: a speech module, logic unit module and object detection module. The computer vision component is boxed in blue.}
\label{oliv-system}
\vspace{-0.1in}
\end{figure*}

\vspace{-0.06in}
The proposed OLIV system~\cite{Wang_OLIV_2018}, shown in Figure \ref{oliv-system} is designed to aid individuals with visual impairment to locate displaced items in an indoor environment, and is comprised of three main components: i) a speech module, ii) an object detection module, and iii) a logic unit module.

When a user wishes to locate a displaced item, he or she issues a verbal query to OLIV.  The speech module, which utilizes an off-the-shelf commercial smart assistant, interprets the user's verbal query and sends a parsed query to the logic unit module. The logic unit module triggers a snapshot of the scene using a camera, which is received by the object detection module. The object detection module identifies all objects of interest and their corresponding object labels and locations. The object detection module then feeds that back to the logic unit module.  Based on the user's intent along with the object type and location information, the logic unit module calculates the relative location of the queried object to the other objects in the scene and constructs a semantic description providing directions for the user to locate the object. Finally, this semantic description is communicated to the speech module to be verbally conveyed back to the user. By providing descriptions of where an object is, the user's mental map can be updated, increasing their independence. Thus, it can be seen that the design of fast, but accurate on-device object detection is critical to the success of OLIV as an assistive tool.

\section{Micro-architecture design exploration}
\vspace{-0.06in}
Inspired by the small and fast architecture of MobileNetV2-SSD~\cite{ssd,mobilenetv2}, this study is motivated to take it one step further by accounting for the specialized application of indoor object detection for individuals with visual impairment required for OLIV, which has a smaller number of target classes. Given the reduced number of classes, an exploration of hyperparameters is conducted to find a network that offers the best trade-offs between performance, speed and size for on-device object detection.

\subsection{Architectural design hyperparameters}
MobileNetV2 includes two additional hyperparameters, width and resolution multiplier. The width multiplier, $\alpha$, affects the width of the network uniformly at each layer by multiplying the number of input channels and output channels by $\alpha$. The resolution multiplier, $\rho$, changes the representation by multiplying each feature map by $\rho$~\cite{mobilenetv2}. Although reducing the width and resolution makes the model smaller and faster, accuracy decreases.

\subsection{Architectural design optimization}
To address the trade-off between accuracy, speed and size, the micro-architecture exploration approach is formulated as a numerical optimization problem, with the goal of finding the set of width and resolution hyperparameter pairs, defined as $\underline{\theta}=\{\alpha,\rho\}$, that maximize a modified NetScore. NetScore is a quantitative assessment of the balance between accuracy, computational complexity and network architecture complexity of a deep neural network~\cite{wong_netscore}. The modified NetScore function is defined as:
\begin{equation}
\label{mod-netscore}
\Omega(\mathcal{N}(\underline{\theta})) = 20\log\Big(\frac{a(\mathcal{N}(\underline{\theta}))^\kappa}{p(\mathcal{N}(\underline{\theta}))^\beta r(\mathcal{N}(\underline{\theta}))^\gamma}\Big)
\end{equation}
where $a(\mathcal{N}(\underline{\theta}))$ is the accuracy of the network, $p(\mathcal{N}(\underline{\theta}))$ is the number of parameters in the network in millions, $r(\mathcal{N}(\underline{\theta}))$ is the CPU running time in seconds, and $\kappa$, $\beta$, $\gamma$ control the the influence of accuracy, architectural complexity and computational complexity, respectively. While architectural and computational complexity are important for this application, accuracy remains the most important metric in the objective function since accuracy is directly correlated to the potential usefulness and benefits to the user's current lifestyle. To reflect the requirements, the weightings for accuracy, speed and size are set to $\kappa=1$, $\beta=0.45$ and $\gamma=0.2$ respectively.

The formal objective function used to determine the most optimal hyperparameters for this application is defined as:
\begin{equation}
\label{obj-func}
    \underline{\hat{\theta}}= \argmax_{\underline{\theta}} \Omega(\mathcal{N}(\underline{\theta}))
\end{equation}
where $\underline{\hat{\theta}}$ is the width and resolution hyperparameter pair that maximizes the modified NetScore, $\Omega(\mathcal{N}(\underline{\theta}))$.

\section{Results and discussion}
\vspace{-0.06in}
 For this study, MSCOCO-OLIV dataset, a 21 class subset of the labeled COCO2017~\cite{coco} train and val images that contain indoor objects, is used for training and evaluation. Different models are trained end-to-end for 800k steps using an initial learning rate of 0.004, decaying by a factor of 0.95 every 200k steps. Each model's backbone architecture is initialized with the pretrained weights from the closest MobileNetV2 classification model trained on ImageNet~\cite{pretrained-mnetv2}.

\begin{figure}[t]
\begin{center}
\includegraphics[width=0.8\linewidth]{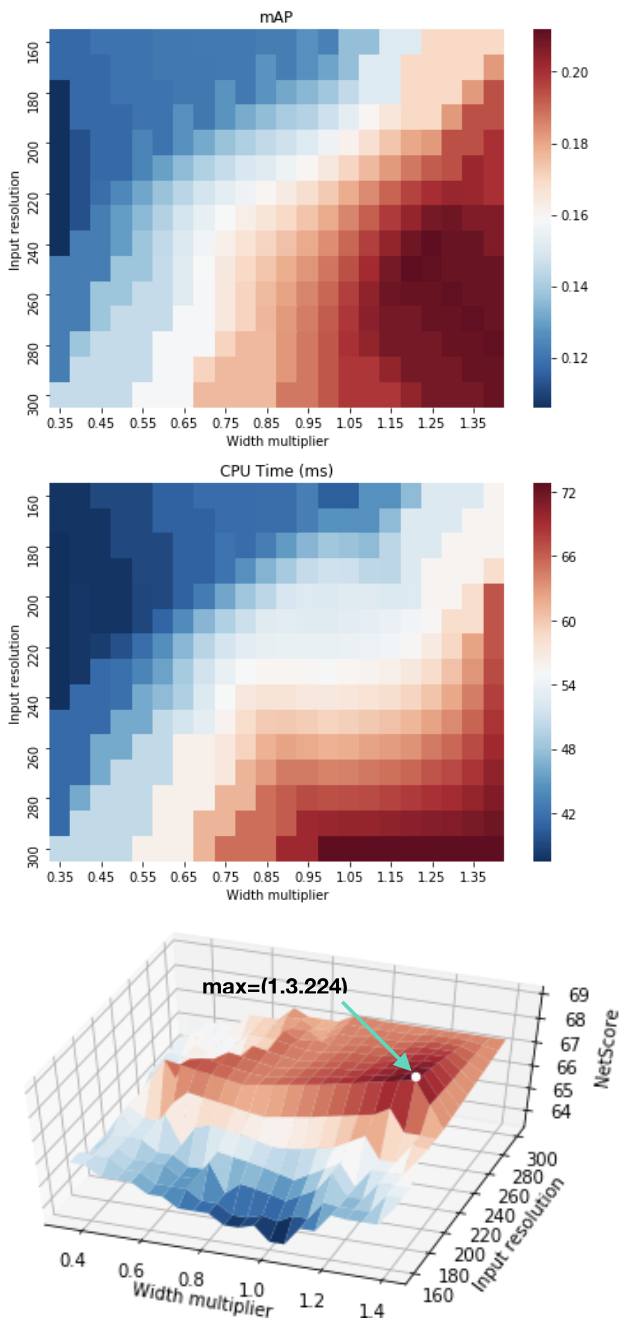}
\end{center}
   \caption{Effect of width multiplier and input resolution on mAP (top) and CPU time (middle), and modified NetScore results shown as a surface plot (bottom). Arrow indicate optimal model.}
\label{results}
\end{figure}

Figure \ref{results} offers an in-depth insight into the effects of the the width multiplier and input resolution on overall mAP and CPU running time. These results show that in general, as the width multiplier and input resolution increase, the mAP and CPU run-time also increase. However, there is a certain point where there are diminishing returns in the increase of mAP. Based on the mAP plot, there is little increase in mAP after the width multiplier reaches 1.15 and input resolution reaches 220. However, in the CPU time plot, run-time continues to increase linearly as the width multiplier and input resolution increase.

Although the previous analysis results offer insight on the effects of width and resolution hyperparameters, those results do not give a quantitative answer as to which model offers the best trade-offs for the case of assistive devices for the visually impaired. As such, we employ the architectural design optimization strategy mentioned in Section 3.2. To take a deeper look into this architecture design optimization process, Figure \ref{results} illustrates how the modified NetScore changes depending on the width multiplier and input resolution. Based on Figure \ref{results}, the model with hyperparameter pairs, width multiplier of 1.3 and input resolution of 224, achieves the highest score of 68.7, indicating that this model offers the best balanced trade-offs for OLIV.

In this paper, we investigated a micro-architecture exploration approach to create an object detection network specifically for OLIV. Based on experimental results, this approach resulted in a well-suited compact object detection model that offers a balanced trade-off between accuracy, speed and size. Moreover, this approach has the potential to be applied to other assistive devices to offer users a cost-efficient and secure solution.

{\small
\bibliographystyle{ieee}
\bibliography{References}
}

\end{document}